\documentclass[letterpaper, 10 pt, conference]{ieeeconf}  % Comment this line out if you need a4paper

\IEEEoverridecommandlockouts                              % This command is only needed if 
\usepackage{todonotes}
\usepackage{support-caption}
\usepackage{subcaption}
\usepackage{amsmath}
\usepackage{comment}
\usepackage{multirow}

\usepackage{amssymb}
\usepackage{booktabs}

\usepackage[normalem]{ulem}

\usepackage{titlesec}
\titleformat{\paragraph}[runin]
{\normalfont\normalsize\bfseries}{}{1em}{}
\titlespacing{\paragraph}
{0pt}{3.25ex plus 1ex minus .2ex}{1em}

\usepackage{scalerel}
\usepackage{tikz}
\usetikzlibrary{svg.path}

\definecolor{orcidlogocol}{HTML}{A6CE39}
\tikzset{
  orcidlogo/.pic={
    \fill[orcidlogocol] svg{M256,128c0,70.7-57.3,128-128,128C57.3,256,0,198.7,0,128C0,57.3,57.3,0,128,0C198.7,0,256,57.3,256,128z};
    \fill[white] svg{M86.3,186.2H70.9V79.1h15.4v48.4V186.2z}
                 svg{M108.9,79.1h41.6c39.6,0,57,28.3,57,53.6c0,27.5-21.5,53.6-56.8,53.6h-41.8V79.1z M124.3,172.4h24.5c34.9,0,42.9-26.5,42.9-39.7c0-21.5-13.7-39.7-43.7-39.7h-23.7V172.4z}
                 svg{M88.7,56.8c0,5.5-4.5,10.1-10.1,10.1c-5.6,0-10.1-4.6-10.1-10.1c0-5.6,4.5-10.1,10.1-10.1C84.2,46.7,88.7,51.3,88.7,56.8z};
  }
}

\newcommand\orcidicon[1]{\href{https://orcid.org/#1}{\mbox{\scalerel*{
\begin{tikzpicture}[yscale=-1,transform shape]
\pic{orcidlogo};
\end{tikzpicture}
}{|}}}}

\usepackage{hyperref}
\usepackage{float}
\title{\huge Tensor-Based Self-Calibration of Cameras via the TrifocalCalib Method}

\author{Gregory Schroeder$^{1, 2}$\orcidicon{0009-0005-7340-1715} \textit{Member, IEEE}
, Mohamed Sabry$^{1}$\orcidicon{0000-0002-9721-6291} \textit{Member, IEEE}, \\ 
and Cristina Olaverri-Monreal$^{1}$\orcidicon{0000-0002-5211-3598} \textit{Senior Member, IEEE}%
\thanks{$^1$ Johannes Kepler University Linz, Austria, Department Intelligent Transport Systems
\texttt{\{gregory.schroeder, cristina.olaverri-monreal\}@jku.at}}%
\thanks{$^{2}$Intelligent Systems Functions Department, IAV GmbH, Berlin, Germany 
{\tt\footnotesize gregory.schroeder@iav.de}}%
}

\begin{document}
\maketitle
\thispagestyle{empty}
\pagestyle{empty}

\setlength{\abovecaptionskip}{4pt}   % Space between Figure and caption

%%%%%%%%%%%%%%%%%%%%%%%%%%%%%%%%%%%%%%%%%%%%%%%%%%%%%%%%%%%%%%%%%%%%%%%%%%%%%%%%

\begin{abstract}

Estimating camera intrinsic parameters without prior scene knowledge is a fundamental challenge in computer vision. 
This capability is particularly important for applications such as autonomous driving and vehicle platooning, where pre-calibrated setups are impractical and real-time adaptability is necessary.
%
% In this work, we introduce equations derived from the calibrated trifocal tensor for projective camera self-calibration. 
To advance the state-of-the-art, we present a set of equations based on the calibrated trifocal tensor, enabling projective camera self-calibration from minimal image data.
Our method, termed \textit{TrifocalCalib}, significantly improves accuracy and robustness compared to both recent learning-based and classical approaches.
Unlike many existing techniques, our approach requires no calibration target, imposes no constraints on camera motion, and simultaneously estimates both focal length and principal point. 
Evaluations in both procedurally generated synthetic environments and structured dataset-based scenarios demonstrate the effectiveness of our approach.
%We make the code publicly available at: 
%\scriptsize\url{https://gitlab.com/intelligent-transportation-systems/pdrive/projective_camera_selfcalibration}
%
% We make the code publicly available.
To support reproducibility, we make the code publicly available.
\footnote[3]{\scriptsize\url{https://gitlab.com/intelligent-transportation-systems/pdrive/projective_camera_selfcalibration}}

\end{abstract}

%\vspace{-5pt}
\section{Introduction}
\label{sec:introduction}

Camera self-calibration, also known as auto-calibration, aims to estimate the intrinsic parameters of a camera directly from an image sequence, without requiring a predefined calibration object or prior knowledge of the scene. This eliminates the need for controlled environments or additional information such as checkerboards, parallel lines, or vanishing points, making it highly relevant for real-world applications, such as autonomous and connected vehicles,
where such constraints are impractical. Accurate intrinsic parameter estimation is crucial for many computer vision tasks, including 3D reconstruction, structure-from-motion, vehicle platooning and augmented reality.

% Traditional self-calibration methods often rely on the Fundamental Matrix, which encodes epipolar geometry between image pairs. However, these methods are often sensitive to noise, require specific camera motions, or exhibit instability in practical scenarios. 

Many existing methods simplify this task by performing only partial self-calibration, estimating the focal length while assuming the principal point is fixed at the image center. Although this reduces complexity, it is well established that the principal point often deviates significantly from the image center, limiting calibration accuracy. In contrast, our work addresses complete self-calibration, jointly estimating both focal length and principal point. 
We introduce equations derived from the calibrated Trifocal Tensor that enhance the accuracy and robustness of projective self-calibration. Our method does not require constraints on camera motion and achieves significantly better results compared to Fundamental Matrix-based approaches and recent learning-based approaches. 
To validate our approach, we conduct experiments in both procedurally generated synthetic environments and structured dataset-based scenarios, ensuring a comprehensive evaluation of our method's effectiveness.
%
% We assume the images to be undistorted. The distortion parameters can be estimated independent of the projective camera parameters~\cite{devernay2001straight, schroeder2025clearlines} 

\vspace{-10pt}
\paragraph{Contributions:}
\begin{enumerate}
    \item We propose a novel calibration method based on the algebraic constraints that characterize a metric trifocal tensor.
    \item We conduct a detailed evaluation of the proposed formulation on synthetic data with known ground truth.
    \item We integrate the trifocal constraints into a hybrid calibration framework that combines deep learning-based feature extraction and matching with geometric reasoning.
    \item We validate the method on the BlendedMVS dataset~\cite{2020_blendedmvs} and demonstrate superior performance compared to state-of-the-art approaches.
    \item We publicly release the code to facilitate reproducibility and further research. \footnotemark[3]
    % \footnote{\scriptsize \url{https://gitlab.com/intelligent-transportation-systems/pdrive/projective_camera_selfcalibration}}
\end{enumerate}

\vspace{-10pt}
\paragraph{Paper Outline:}
Section~\ref{sec:related_work} surveys existing literature. 
Section~\ref{sec:methodology} details the mathematical basis of the used approaches.
% details the theoretical foundations of our proposed method
%
Section~\ref{sec:experiments_synthetic} and~\ref{sec:experiments_image} provide an overview of the conducted experiments. 
% describes the experimental setup, including data generation and evaluation metrics
%
Section~\ref{sec:results} reports the results.
Finally, Section~\ref{sec:conclusion} highlights key contributions and conclusions.

\section{Related Work} 
\label{sec:related_work}

% Overview of existing methods for camera parameter estimation.

% Comparison with classical and modern approaches (e.g., fundamental matrix, trifocal tensor, bundle adjustment).

% \todo[inline]{Include 2024 paper 'Minimal Perspective Autocalibration' - Andrea Porfiri Dal Cin}
% Recently, Porfiri et al.~\cite{2024_BA_minimal} focus on minimal problem formulations and efficient solvers for two and three views. The camera intrinsics and 3D point depths are jointly estimated.

% \todo[inline]{include: Camera Self-Calibration Using the Singular Value Decomposition of the Fundamental Matrix - Manolis I.A. Lourakis and Rachid Deriche}

\vspace{-2pt}
\paragraph{Classical Approaches}
Classical approaches to intrinsic camera self-calibration rely on geometric multi-view constraints. 
% Early techniques utilized the Fundamental Matrix, which encapsulates epipolar geometry between image pairs. 
Initial methods introduced Kruppa’s equations derived from rigidity constraints~\cite{1992_faugeras_self_calibration_Theory_and_experiments}. A later simplification exploited the essential matrix's intrinsic properties~\cite{1999_A_simple_technique_for_self_calibration}, yet fundamental matrix-based methods remain sensitive to noise and often demand specific camera motions.
Moreover, the resulting objective functions typically display numerous local minima, 
% why subsequent work focuses on global optimization approaches, such as 
which is why subsequent work has focused on global optimization approaches~\cite{2008_ze_fundamnetal_particle_swarm, 2004_whitehead_fundamental_evolutionary}.
% such as evolutionary optimization~\cite{2004_whitehead_fundamental_evolutionary}, particle swarm optimization~\cite{2008_ze_fundamnetal_particle_swarm} or dynamic hill climbing~\cite{2002_roth_fundamental_dynamic_hill_climbing}. 
However, these techniques are computationally expensive and offer no guarantee of finding the global optimum. 
%
% Trifocal Tensor
% To address these limitations, the Trifocal Tensor was explored as it provides richer geometric constraints over three views. Early work demonstrated improved robustness~\cite{1996_armstrong_trifocal,1998_faugeras_trifocal}, but remained limited to planar camera motions.
% 
% Another classical strategy is Bundle Adjustment (BA), a nonlinear optimization technique that jointly refines camera parameters, pose transformations, and 3D scene structures by minimizing reprojection error~\cite{triggs2000bundle}. 
A classical strategy that jointly refines camera parameters, pose transformations, and 3D scene structures by minimizing reprojection error is Bundle Adjustment (BA), a nonlinear optimization technique~\cite{triggs2000bundle}. 
While BA is a often a crucial step in achieving high-accuracy, it is generally not considered a self-calibration method on its own. Instead, it is commonly used as a post-processing refinement step, requiring good initial estimates of camera intrinsics, extrinsics, and 3D points to converge to an optimal solution. 
% Without reasonable initialization, BA can fail or converge to local minima, limiting its applicability as a standalone calibration method. 
%
% Recently, Porfiri et al.~\cite{2024_BA_minimal} focus on minimal problem formulations and efficient solvers for two and three views. The camera intrinsics and 3D point depths are jointly estimated.
A comprehensive introduction to the Fundamental Matrix, Trifocal Tensor, Bundle Adjustment, and classical computer vision methods can be found in~\cite{2003_hartley_zisserman}.

\vspace{-10pt}
\paragraph{Deep Learning-Based Methods}
Recent advancements in deep learning have led to novel self-calibration approaches that bypass explicit geometric computations. 
DeepPTZ~\cite{2020_zhang_deepptz} introduced a method that automatically estimates focal length and distortion parameters using a dual-Siamese network structure. 
% These approaches offer the potential for real-time calibration and adaptability to diverse imaging conditions.
%
PerspectiveFields~\cite{jin2023perspective} models local perspective properties of an image through per-pixel up vectors and latitude values. This representation allows for robust estimation of camera parameters and has applications in image compositing.
Self-supervised techniques have emerged to perform online camera calibration without the need for labeled data or controlled environments. 
In~\cite{2022_fang_self_supervised} a self-supervised method was developed, that learns intrinsic parameters from raw video sequences, achieving sub-pixel reprojection error across various camera models. 
%
% Similarly, a framework for self-supervised online calibration targeting automated driving and parking was presented, demonstrating its applicability in dynamic real-world conditions~\cite{2023_hogan_deep}.
The authors in~\cite{2023_hogan_deep} applied a similar technique to automated driving and parking, proving its usefulness in real-world dynamic scenarios.
%
% Additionally, practical implementations, such as the DeepCalib system~\cite{2018_bogdan_deepcalib}, utilize deep learning to estimate intrinsic parameters from single images, demonstrating the feasibility of integrating calibration into real-world applications.
Moreover, approaches like DeepCalib~\cite{2018_bogdan_deepcalib} utilize deep neural networks to infer intrinsic parameters from individual images, underscoring the viability of calibration in operational environments.

\vspace{-10pt}
\paragraph{Hybrid Approaches}
Hybrid methods combining traditional geometric techniques with modern learning-based approaches have also been explored. 
For example, DroidCalib~\cite{2023_hagemann_learning_based_with_code} proposed a deep learning approach that integrates a differentiable self-calibrating bundle adjustment layer, enabling the estimation of camera intrinsics within a neural network framework. 
They also report results for a hybrid enhanced solution of the classical COLMAP~\cite{2016_colmap_schoenberger} pipeline, by integrating traditional geometric optimization with learned feature detectors and matchers such as SuperPoint~\cite{2018_superpoint} and SuperGlue~\cite{2020_superglue}. 
SceneCalib~\cite{2023_scenecalib} presents a method for simultaneous self-calibration of extrinsic and intrinsic parameters in systems containing multiple cameras and lidar sensors, addressing the need for accurate sensor fusion in 3D perception tasks. 
The authors in~\cite{2021_Lee_ICCV} integrate a Transformer model with geometric priors in the form of line-segment features. By combining deep learning with explicit geometry, the approach provides strong calibration performance, especially in scenes with dominant line structures. The model predicts multiple intrinsic parameters requiring either manually provided or automatically detected line features.
\section{Methodology} 
\label{sec:methodology}

% Ground Truth and Perturbation: Explanation of how the true parameters are set and how they are perturbed.

% Error Metric: Definition of the mean relative distance metric.

% Estimation Methods:
    % Fundamental Matrix approach
    % Trifocal Tensor approach
    % Bundle Adjustment

In this section, we introduce our method for self-calibrating the intrinsic camera parameters - focal lengths $(f_x, f_y)$ and principal point $(c_x, c_y)$ - from multiple uncalibrated views using constraints derived from the calibrated trifocal tensor. 
We begin by briefly recalling the definition of the trifocal tensor and its role in multiview geometry. For a detailed treatment of projective and multiview geometry concepts, including the trifocal tensor, see~\cite{2003_hartley_zisserman}.

\paragraph{The Trifocal Tensor.}
The trifocal tensor describes the geometric relationship among three views and is central to three-view geometry. It encodes the relative camera motion and epipolar geometry without requiring knowledge of scene structure. In the calibrated case, the trifocal tensor inherits additional algebraic constraints due to the known intrinsic parameters. 
Given three projection matrices $P_1 = [I \;|\; 0]$, $P_2 = [R_2 \;|\; t_2]$, and $P_3 = [R_3 \;|\; t_3]$, the calibrated trifocal tensor $\hat{T}$ has correlation slices $\hat{T}_k$ defined as:
\[
\hat{T}_k = R_2 e_k t_3^\top - t_2 e_k^\top R_3^\top
\]
where $e_k$ is the $k$-th basis vector.

\vspace{-5pt}
\paragraph{Polynomial Constraints for Self-Calibration.}
The space of calibrated trifocal tensors was characterized in~\cite{2017_martyushev_one_some_properties_of_calibrated_trifocal_tensors} by a set of necessary algebraic constraints. Among these are 15 quartic polynomial conditions derived from matrix trace and rank relationships. For real-valued trifocal tensors, these constraints are not only necessary but also sufficient. They define a nonlinear system in the entries of the calibrated trifocal tensor and, by extension, impose constraints on the intrinsic camera parameters.
% We summarize the 15 quartic constraints derived in~\cite{2017_martyushev_one_some_properties_of_calibrated_trifocal_tensors} as follows.

Let the symmetric matrices be defined as:
\[
U_k = \hat{T}_k \hat{T}_k^\top, \quad
V_k = \hat{T}_k \hat{T}_{k+1}^\top + \hat{T}_{k+1} \hat{T}_k^\top
\]
with cyclic indexing $k+1$ modulo 3.
Define:
\[
\psi(X,Y) = \mathrm{tr}(X)\mathrm{tr}(Y) - 2\,\mathrm{tr}(XY)
\]

Then, the first 9 quartic constraints are:
\begin{align*}
\psi(U_3 - U_1, U_3 - U_1) - \psi(V_3, V_3) &= 0 \\
\psi(U_3 - U_1, V_1) + \psi(V_2, V_3) &= 0 \\
\psi(U_1 - U_2, V_1) &= 0
\end{align*}
along with 6 more obtained by cyclic permutation of indices.

The remaining 6 quartic constraints are:
\begin{align*}
\mathrm{tr}(U_2)^2 - \mathrm{tr}(V_3)^2 - \mathrm{tr}(U_2^2\! -\! V_3^2\! +\! (U_3\! -\! U_1)^2) &= 0 \\
\mathrm{tr}(V_2)\mathrm{tr}(U_1\! -\! 2U_2\! -\! U_3) - \mathrm{tr}(V_1)\mathrm{tr}(V_3) + 2\,\mathrm{tr}(V_2 U_2) &= 0
\end{align*}

\begin{figure}[ht]
\centering
\includegraphics[width=0.7\linewidth]{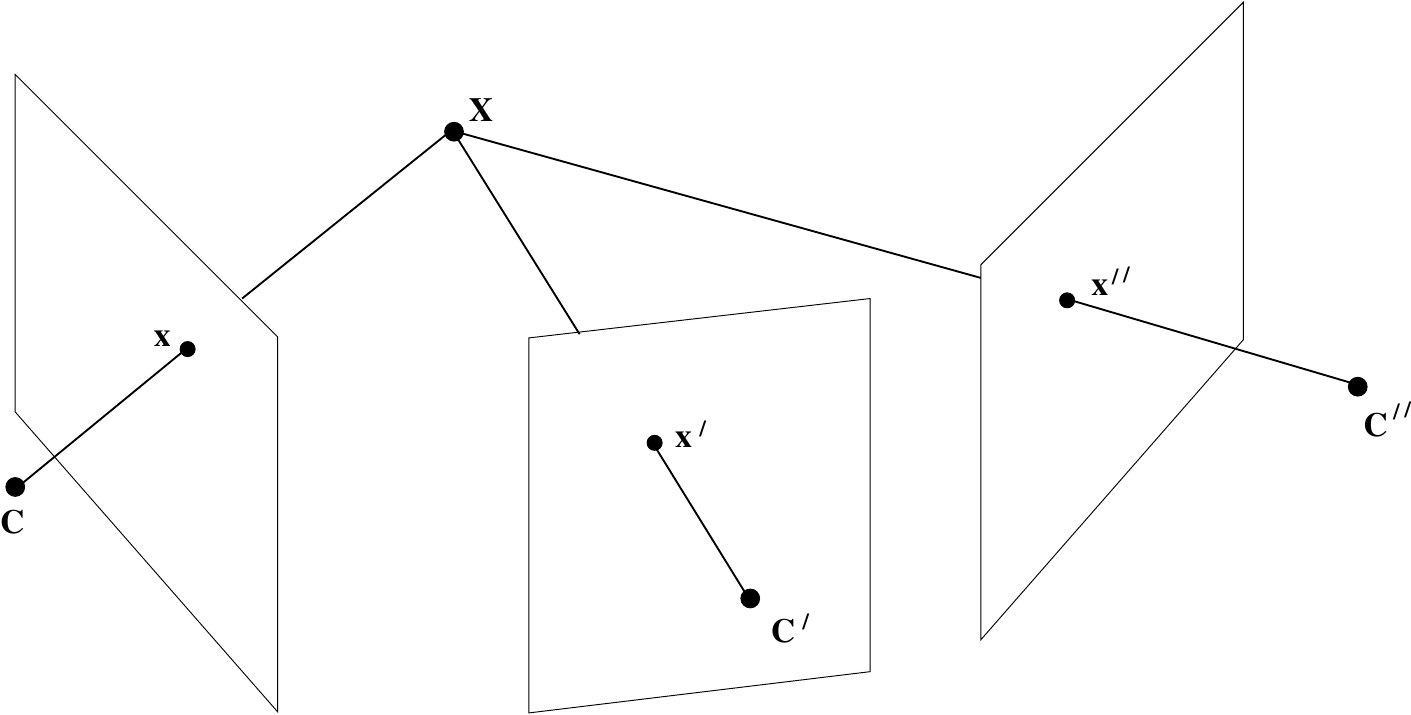}
% \caption{Three views point correspondence geometry 
\caption{Visualization of the trifocal geometry~\cite{2003_hartley_zisserman}. A 3D point \( X \) is projected into three views as image points \( x \), \( x' \), and \( x'' \). These correspondences form the basis for estimating the trifocal tensor.}
\label{fig:3viewfigs}
\end{figure}

\vspace{-5pt}
\paragraph{Parameter Estimation on Synthetic Data.}
In the synthetic setting, we assume known ground truth 3D points and exact camera poses and intrinsics. This allows the generation of ideal correspondences across three views. For realistic evaluation, noise is added to the 2D projections; details of the data setup are provided in the experimental section~\ref{sec:experiments_synthetic}.
%
% Given the correspondences, we compute a linear estimate of the trifocal tensor using the standard algorithm based on solving a homogeneous linear system~\cite{2003_hartley_zisserman}. This method is computationally efficient and provides a unique, globally optimal solution in the least-squares sense.
Given a set of corresponding image points across three views (see Figure~\ref{fig:3viewfigs}), we compute an estimate of the trifocal tensor using the standard linear algorithm~\cite{2003_hartley_zisserman}, which formulates a homogeneous linear system based on multilinear constraints. Specifically, the trifocal tensor \( T_i \) relates point correspondences \( x \in \mathbb{P}^2 \), \( x' \in \mathbb{P}^2 \), and \( x'' \in \mathbb{P}^2 \) across three views via the following equation:

\begin{equation*}
	[x']_{\times} \left( \sum_{i=1}^{3} x^i T_i \right) [x'']_{\times} = 0_{3\times3}
	\label{equ:tensor_point_correspondence}
\end{equation*}

Each point triple provides a set of linear constraints on the tensor slices \( T_i \in \mathbb{R}^{3 \times 3} \). Stacking these equations over all correspondences results in an overdetermined linear system, which we solve in the least-squares sense using singular value decomposition (SVD). This yields a unique, globally optimal estimate of the trifocal tensor under the linear model.

We then estimate the intrinsic parameters $(f_x, f_y, c_x, c_y)$ by minimizing the violation of the 15 quartic constraints. The optimization is formulated as:

\[
\min_{\mathbf{K}} \sum_{i=1}^{15} \left( \phi_i(\hat{T}_1(\mathbf{K}), \hat{T}_2(\mathbf{K}), \hat{T}_3(\mathbf{K})) \right)^2
\]

where $\mathbf{K}$ is the intrinsic matrix, and $\phi_i$ denotes the $i$-th quartic polynomial constraint from~\cite{2017_martyushev_one_some_properties_of_calibrated_trifocal_tensors}.

\vspace{-5pt}
\paragraph{Parameter Estimation on Image Data.}
In contrast to the synthetic setting, estimating camera intrinsics from image data involves several additional steps and challenges. In particular, feature correspondences across three views must be extracted and matched.
To address this, we leverage deep learning-based feature detectors and matchers. Specifically, we use SuperPoint~\cite{2018_superpoint} for keypoint detection and LightGlue~\cite{2023_lightglue} for matching. While these tools provide high-quality correspondences, they also introduce outliers and noises.
We apply a RANSAC-based~\cite{ransac} estimation framework to handle outliers.
The performance of RANSAC is sensitive to the inlier threshold, which is often ambiguously treated in the literature. We therefore use MSAC~\cite{2000-msac-mlesac}, a variant that improves robustness by reducing sensitivity to the threshold and providing a more stable scoring function for model selection.
The MSAC-score is defined as:

\begin{equation}
\text{MSAC-score} = 1 - \frac{1}{N} \sum_{i=1}^{N} \min\left(\frac{r_i}{\tau}, 1\right)
\label{eq:msac_score}
\end{equation}

Here, \( r_i \) denotes the individual residual values, \( N \) is the number of residuals, and \( \tau \) a predefined inlier threshold. Residuals below the threshold contribute proportionally, while those exceeding the threshold are uniformly capped. This formulation promotes solutions with predominantly small residuals while remaining robust to outliers.

Despite these differences in preprocessing and outlier handling, the calibration step remains the same as in the synthetic case: we optimize the intrinsic parameters $(f_x, f_y, c_x, c_y)$ by minimizing the violation of the 15 quartic constraints of the calibrated trifocal tensor.

% \paragraph{Comparison with Fundamental Matrix-Based Methods.}
% Our method draws a parallel to Mendonça and Cipolla’s approach~\cite{mendonca1999simple}, which estimates intrinsic parameters by transforming the fundamental matrix into an essential matrix and enforcing singular value constraints. However, while that technique is limited to pairwise views and two constraints per pair, our approach exploits richer geometric constraints available in the trifocal tensor, enabling more robust and accurate estimation from image triplets.

\section{Experiments - Synthetic Data} 
\label{sec:experiments_synthetic}

Synthetic data allows precise control over parameters such as noise, outlier ratio, and initial perturbations of the camera intrinsics. This makes it ideal for evaluating the robustness and accuracy of self-calibration methods in isolation from confounding real-world effects.

\subsection{Metrics}
\label{ssec:Metrics}
To evaluate the accuracy of the estimated intrinsic camera parameters, we compute the mean relative error with respect to the ground truth. Specifically, we compare the estimated focal lengths $(f_x, f_y)$ and principal point coordinates $(c_x, c_y)$ against their known ground truth values. 

For each parameter, we calculate the relative distance:
\[
\text{rel\_dist}_{p} = \left| \frac{p_\text{gt} - p_\text{est}}{p_\text{gt}} \right|,\quad p \in \{f_x, f_y, c_x, c_y\}
\]

The final metric is the mean of the four relative distances:
\begin{align*}
\text{mean\_error} = \frac{1}{4} (&\text{rel\_dist}(f_x) + \text{rel\_dist}(f_y) \\
                                &+ \text{rel\_dist}(c_x) + \text{rel\_dist}(c_y))
\end{align*}

% This aggregated metric reflects the overall deviation of the estimated intrinsics from the ground truth and is reported for all baseline and proposed methods.
This aggregated metric quantifying the overall deviation of the estimated intrinsic parameters from the ground truth is computed for all baseline and proposed self-calibration methods

\subsection{Experimental setup} 

\begin{figure}[h]
    \centering
    \includegraphics[width=0.85\linewidth]{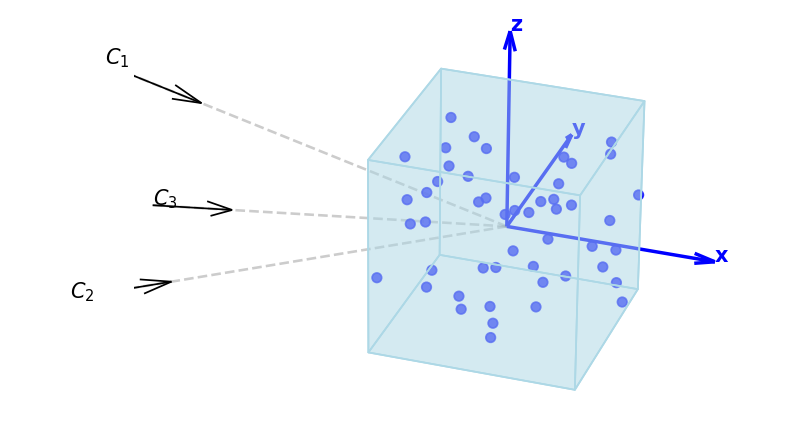}
    \caption{Visualization of the experimental setup for the synthetic data. Three cameras ($C_1$, $C_2$, $C_3$) in general position facing the coordinate origin where 3D-points are generated within the boundaries of a cube.}
    \label{fig:synthetic_data_setup}
\end{figure}

A simplified visualization of the synthetic data setup is shown in Figure~\ref{fig:synthetic_data_setup}.  
A set of 3D points is randomly generated within the boundaries of a cube centered at the coordinate origin.  
Three cameras, $C_1$, $C_2$, and $C_3$, are placed in general position and oriented toward the coordinate origin.  
They share identical intrinsic parameters and capture images of the same size.  
The 3D points are projected onto their image planes using the standard perspective projection model:

\begin{equation}
\mathbf{x} = K [R \ | \ \mathbf{t}] \mathbf{X}
\end{equation}

where $\mathbf{X} \in R^3$ are the 3D points, $[R | \mathbf{t}]$ the extrinsic parameters (rotation and translation), $\mathbf{x} \in R^2$ the corresponding 2D projections, and $K$ the intrinsic camera matrix.

To simulate real-world imperfections, uniformly distributed noise is added to the projected image points.  
The exact number of 3D points and the magnitude of the image noise vary across experiments and are specified for each result.
In addition, the ground truth intrinsic parameters are perturbed to generate an initial guess for self-calibration. Each parameter is randomly sampled from a uniform distribution within $\pm 5\%$ of its ground truth value. 
% This initialization ensures a realistic, inaccurate prior.
This initialization provides a realistic yet deliberately inaccurate prior estimate to test the robustness of the proposed method.

\subsection{Baseline Approaches}
Among the methods considered, only the classical approach presented in~\cite{1999_A_simple_technique_for_self_calibration} can be directly applied to synthetic data, as it can operate purely on sparse point correspondences.
In contrast, learning-based and hybrid approaches are designed to operate on images.
A broader comparison to learning-based and hybrid approaches is conducted on image data in Section~\ref{sec:experiments_image}.

\begin{figure*}[t]%[h]%[H]
\def\heightfigresults{3.3cm}
% First row
\subfloat[n = 0.1]{\includegraphics[height=\heightfigresults,width=5.85cm]{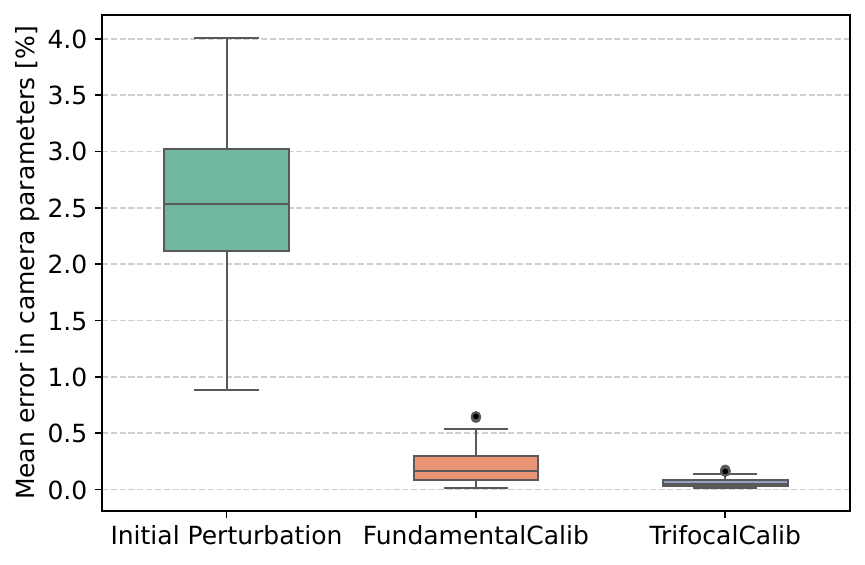}}
\hspace{0.01cm}
\subfloat[n = 0.5]{\includegraphics[height=\heightfigresults,width=5.85cm]{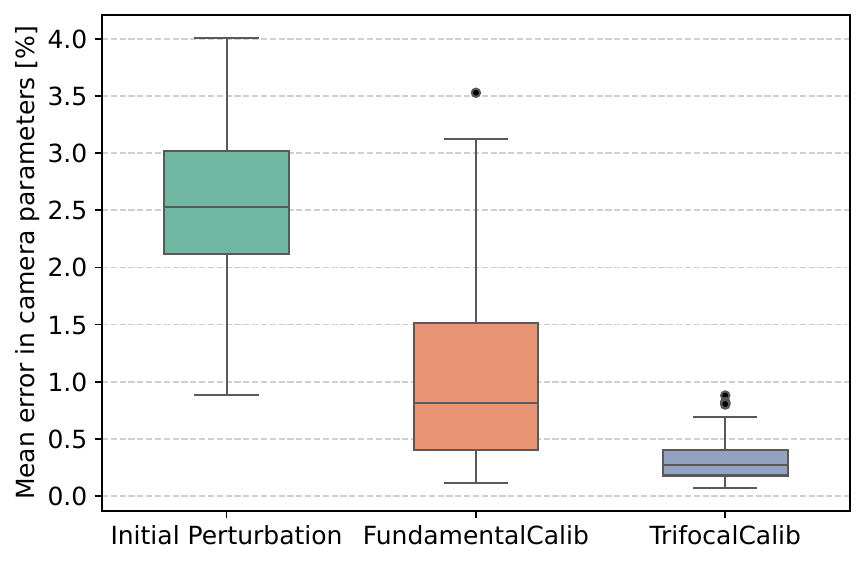}}
\hspace{0.01cm}
\subfloat[n = 1.0]{\includegraphics[height=\heightfigresults,width=5.85cm]{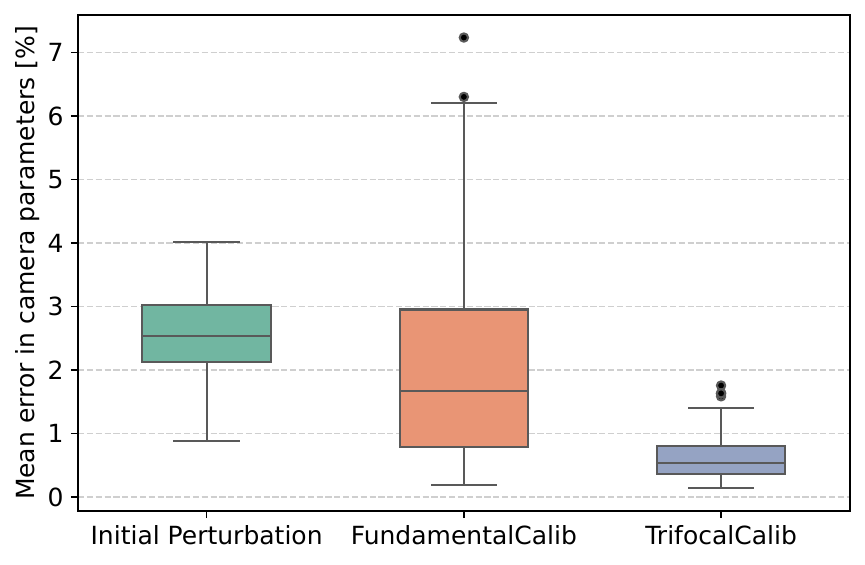}}
\caption{
Boxplots of the mean error, expressed in percent, in estimated intrinsic parameters ($f_x$, $f_y$, $c_x$, $c_y$) under increasing image correspondence noise levels. Noise is uniformly sampled in the range $[-n, +n]$ pixels, with $n \in \{0.1, 0.5, 1.0\}$. The initial perturbation to the ground truth camera parameters is fixed across all experiments, sampled uniformly from $[-5\%, +5\%]$. Each synthetic run uses 500 point correspondences per view triplet. “Initial Perturbation” reflects the pre-calibration error. \textit{FundamentalCalib} and \textit{TrifocalCalib} attempt to recover the original parameters from correspondences. Each box summarizes results over 100 synthetic runs. Lower values indicate better calibration accuracy.
}
\label{fig:Results_synthetic_3D_points}
\end{figure*}

\section{Experiments - Image Data} 
\label{sec:experiments_image}

\paragraph{Metrics}
We adopt the same evaluation metrics as introduced in Section~\ref{sec:experiments_synthetic}, measuring the mean error over all intrinsic parameters.

\vspace{-10pt}
\paragraph{Baseline Approaches}

To this end, we include \textit{PerspectiveFields}~\cite{jin2023perspective}, a recent deep learning-based method for single-image camera calibration, due to its strong performance and public availability.
%
% In contrast, DeepPTZ~\cite{2020_zhang_deepptz} and DeepCalib~\cite{2018_bogdan_deepcalib} were not included as baselines, as they lacked available pretrained weights or did not estimate the principal point, respectively.
%
%
We incorporate the hybrid approach DroidCalib~\cite{2023_hagemann_learning_based_with_code}, which integrates deep learning-based feature extraction and matching with geometric optimization, demonstrating superior performance over prior methods such as SelfSup-Calib~\cite{2022_fang_self_supervised} and the classical version of COLMAP~\cite{2016_colmap_schoenberger} as well as the feature-enhanced version with SuperPoint~\cite{2018_superpoint} and SuperGlue~\cite{2020_superglue}.
% 
% Our method did not take into account the following approaches due to specific limitations:  SceneCalib~\cite{2023_scenecalib} requires LiDAR input and lacks publicly available code, similarly to~\cite{2023_hogan_deep}. The approach in~\cite{2021_Lee_ICCV} assumes visible structural lines in artificial environments and does not estimate the principal point.
%
%
Finally, we also implemented the classical self-calibration technique presented in~\cite{1999_A_simple_technique_for_self_calibration}, which offers a more recent alternative to traditional, Kruppa-equation-based approaches. We refer to it as \textit{FundamentalCalib} in the remainder of this paper.

% \todo[inline]{Implement 4.1 Three-view Auto-Calibration from 'Necessary and Sufficient Polynomial Constraints on Compatible Triplets of Essential Matrices'} 

% \todo[inline]{Solvers of 2 and 3 views~\cite{2024_BA_minimal} has code. Might test it}

% \todo[inline]{Test Bundle Adjustment as postprocessing process, i.e. estimate all parameters with Trifocal or Fundamental and try to improve the results}

\begin{figure}[H]
\def\heightfigresults{2.5cm}
\def\widthtfigresults{4.2cm} % 5.85cm
% First row
\subfloat{\includegraphics[height=\heightfigresults,width=\widthtfigresults]{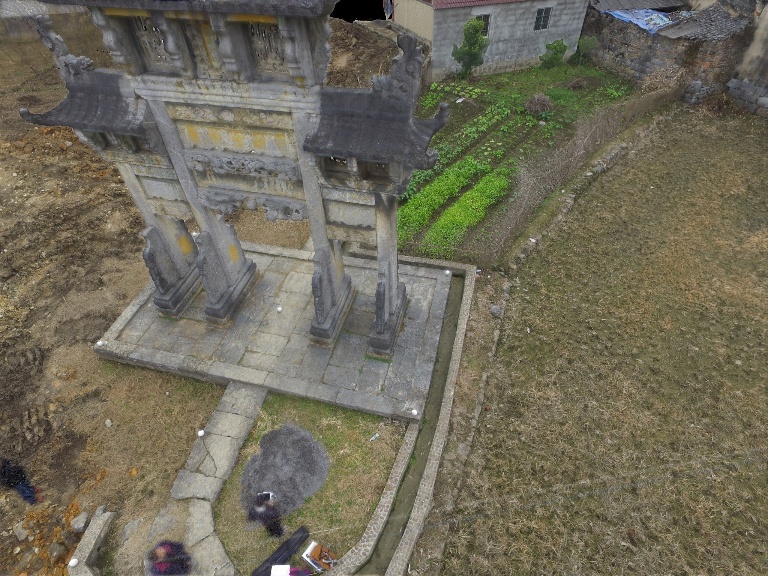}}
\hspace{0.01cm}
\subfloat{\includegraphics[height=\heightfigresults,width=\widthtfigresults]{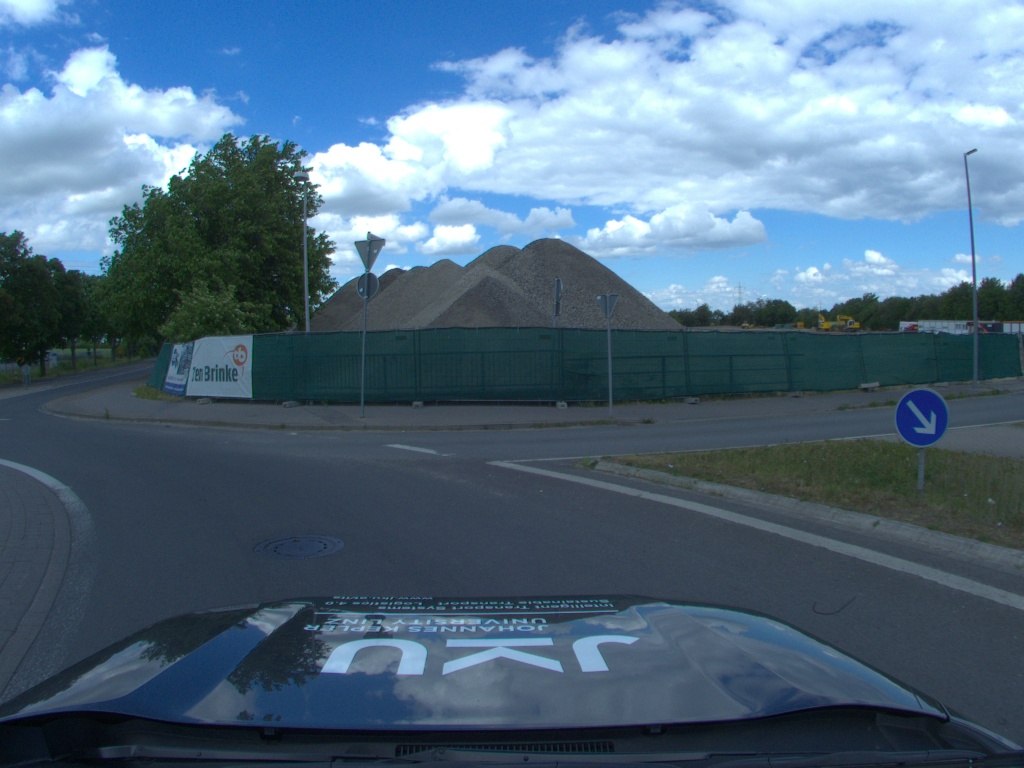}}
% {figures/04_experiments/blendedMVS/00000136_masked.jpg}}
\caption{Example image from the blendedMVS~\cite{2020_blendedmvs} (left) and the IAMCV dataset~\cite{2024-JKU-dataset} (right) used for the experiments.}
\label{fig:blendedMVS}
\end{figure}

\vspace{-10pt}
\paragraph{Experimental Setup}
For the experimental setup, we use the BlendedMVS dataset~\cite{2020_blendedmvs}, which contains photorealistic renderings of real-world scenes with diverse textures, viewpoints, and lighting conditions (see Figure~\ref{fig:blendedMVS} for example images). One of its key advantages is that it provides ground truth camera intrinsics, allowing quantitative evaluation of self-calibration methods. 
%
% To construct an evaluation sequence, we automatically select overlapping views using the provided similarity scores. Starting from image zero, we iteratively select the most similar unused image based on the provided pairwise similarity scores, ensuring view overlap. 
%
Since \textit{PerspectiveFields}~\cite{jin2023perspective} is a single-shot method that estimates camera parameters from individual images, it does not benefit from multi-frame optimization or leverage initial guesses. To evaluate it fairly, we apply it to each image individually and report the average performance over all predictions.
Each of the other methods is evaluated over 100 randomized runs. For each run, synthetic noise is added to the initial camera intrinsics simulating uncertainty in prior calibration. 
We evaluate performance across varying numbers of input images (8, 16, 32) and under different levels of camera parameter noise, perturbing each parameter with a random value uniformly sampled from either the range $[-5\%, +5\%]$ or the more challenging$[-10\%, -5\%] \cup [5\%, 10\%]$ to test robustness under more challenging conditions.

In addition to BlendedMVS~\cite{2020_blendedmvs}, we include a demonstration on a common use case in automotive and robotics: a forward-facing camera mounted on a vehicle moving approximately on a planar surface. 
% 
% While the KITTI dataset~\cite{KittiDataset} would seem a natural choice, it suffers from significant calibration flaws, as analyzed in detail by~\cite{RecalibratingKitti}. In fact, 
%
Unfortunately, all datasets we reviewed rely on average chessboard calibration, which raises concerns about using their provided intrinsics as ground truth for high-precision methods. Despite these limitations, we demonstrate that our approach yields reasonable results in this scenario. 
For this purpose, we use the front-facing camera from the IAMCV dataset~\cite{2024-JKU-dataset} (see Figure~\ref{fig:blendedMVS} for example images). 
%
% This dataset was collected using the Johannes Kepler University, Intelligent Transport Systems research vehicle in real-world driving scenarios that include roundabouts, intersections, country roads, and highways, recorded across diverse locations in Germany.

\vspace{-5pt}
\paragraph{TrifocalCalib Variants}
We evaluate three variants of our method to assess the impact of initialization and optimization:

\begin{itemize}
    \item \textit{TrifocalCalib-Direct}: A straightforward approach that directly minimizes the 15 quartic polynomial constraints~\cite{2017_martyushev_one_some_properties_of_calibrated_trifocal_tensors} across all candidate trifocal tensors, without any explicit outlier rejection.
    
    \item \textit{TrifocalCalib-MSAC}: A RANSAC-based variant that estimates camera parameters for each candidate Trifocal Tensor and selects the one with the highest MSAC-score (see Equation~\ref{eq:msac_score}) across all Trifocal Tensors. 
    % This approach avoids optimization and serves as a fast baseline.
    
    \item \textit{TrifocalCalib-MSAC-Opt}: Our full method, initialized using TrifocalCalib-MSAC and refined by directly minimizing the MSAC-score (see Equation~\ref{eq:msac_score}) over all Trifocal Tensors.
\end{itemize}
%
%These variants help isolate the contributions of RANSAC-style initialization and nonlinear optimization, respectively.

\section{Results} 
\label{sec:results}

\subsection{Synthetic Data}
\label{sec:evaluation_synthetic}

Figure~\ref{fig:Results_synthetic_3D_points} shows the distribution of the mean relative error in the estimated intrinsic camera parameters ($f_x$, $f_y$, $c_x$, $c_y$) across increasing levels of image correspondence noise. 
%The two evaluated approaches include the the classical method (“FundamentalCalib”~\cite{1999_A_simple_technique_for_self_calibration}), and our proposed method (“TrifocalCalib”). 
Noise in the image features is uniformly distributed in $[-n, +n]$ pixels, with $n \in \{0.1, 0.5, 1.0\}$. 
The “Initial Perturbation” baseline reflects the error prior to any self-calibration and sets an upper bound for comparison.
Across all noise levels, \textit{TrifocalCalib} consistently achieves the lowest median error and tighter interquartile ranges, indicating both higher accuracy and robustness. 
% As noise increases, the error increases for all methods, but \textit{TrifocalCalib} exhibits a more gradual degradation compared to \textit{FundamentalCalib}. 
% In particular, for the highest noise level ($\pm1.0$ px), the median error of \textit{TrifocalCalib} remains well below that of the alternative.
%
\textit{FundamentalCalib}, while effective at low noise levels, exhibits higher variance and a more significant performance drop under strong correspondence noise. This reflects the sensitivity of the fundamental matrix-based constraints to noise and the advantage of using richer trifocal geometry.
 
% Both methods improve substantially over this baseline, confirming the utility of geometric constraints in recovering intrinsic parameters from image data alone.
% These results demonstrate that \textit{TrifocalCalib} is not only more accurate but also more stable under increasing feature noise.

\subsection{Image Data - BlendedMVS Dataset}
\label{sec:evaluation_images}

\begin{table*}[t]
\centering
\caption{Intrinsic calibration error for BlendedMVS~\cite{2020_blendedmvs} scenes under varying image counts (8/16/32) and initial camera parameter perturbation ranges 
% $[-5\%, +5\%]$ and $[-10\%,-5\%] \cup [5\%,10\%]$. 
($[0\%, 5\%]$ and $[5\%, 10\%]$). 
Results are reported as mean / median / standard deviation in percent, for each scene and number of input images. Each value is averaged over 100 runs. Lower is better.}
\label{tab:blendedmvs_results}

\begin{tabular}{ll|ccc|ccc}
\toprule
\textbf{Images} & \textbf{Method}
& \multicolumn{3}{c|}{\textbf{Scene1 "5a3ca9cb270f0e3f14d0eddb"}}
& \multicolumn{3}{c}{\textbf{Scene2 "5a4a38dad38c8a075495b5d2"}} \\
& 
& Mean [\%] $\downarrow$ & Median [\%] $\downarrow$ & Std. Dev. [\%] $\downarrow$ 
& Mean [\%] $\downarrow$ & Median [\%] $\downarrow$ & Std. Dev. [\%] $\downarrow$ \\
& 
& (0--5 / 5--10) & (0--5 / 5--10) & (0--5 / 5--10)
& (0--5 / 5--10) & (0--5 / 5--10) & (0--5 / 5--10) \\
\midrule

\multirow{4}{*}{8}
& PerspectiveFields & 15.51 / 15.51 & 15.69 / 15.69 & 7.03 / 7.03 & 17.57 / 17.57 & 16.44 / 16.44 & 10.05 / 10.05 \\
& FundamentalCalib  &  0.43 / 0.43 & 0.43 / 0.43 & 0.01 / 0.01 & 1.61 / 1.61 & 1.61 / 1.61 & 0.02 / 0.01 \\    
& DroidCalib        & 2.24 / 2.34   & 2.23 / 2.24   & 0.17 / 0.66 & 7.82 / 11.26  & 7.41 / 10.88  & 3.43 / 4.16 \\
& TrifocalCalib-Direct & \textbf{0.31 / 0.31} & \textbf{0.29 / 0.28} & 0.12 / 0.01 & \textbf{0.45 / 0.45} & \textbf{0.45 / 0.45} & 0.01 / 0.01 \\
& TrifocalCalib-MSAC & 0.66 / 0.66 & 0.66 / 0.66 & \textbf{2e-5 / 2e-5} & 0.72 / 0.72 & 0.72 / 0.72 & \textbf{3e-6 / 3e-6} \\
& TrifocalCalib-MSAC-Opt & 0.46 / 0.46 & 0.46 / 0.46 & 0.002 / 0.002 & 0.57 / 0.58 & 0.56 / 0.56 & 0.05 / 0.06 \\
\midrule

\multirow{4}{*}{16}
& PerspectiveFields & 17.50 / 17.50 & 17.23 / 17.23 & 6.31 / 6.31 & 13.81 / 13.81 & 9.15 / 9.15 & 9.27 / 9.27 \\
& FundamentalCalib  & 3.37 / 3.37 & 3.37 / 3.37  &  0.01 / 0.005 & 0.36 / 0.35  & 0.35 / 0.35  & 0.02 / 0.02 \\    
& DroidCalib        & 1.23 / 1.99   & 1.23 / 1.28   & 0.10 / 5.67 & 0.39 / 0.42   & 0.34 / 0.37   & 0.24 / 0.24 \\
& TrifocalCalib-Direct & 0.26 / 0.26 &  0.25 / 0.25 & 0.03 / 0.05 & 0.41 / 0.42 & 0.44 / 0.45 & 0.10 / 0.08 \\
& TrifocalCalib-MSAC & 0.66 / 0.66 & 0.66 / 0.66 & \textbf{2e-5 / 2e-5} & 0.72 / 0.72 & 0.72 / 0.72 & \textbf{3e-6 / 3e-6} \\
& TrifocalCalib-MSAC-Opt & \textbf{0.42 / 0.42} & \textbf{0.42 / 0.42} & 0.002 / 0.001 & \textbf{0.25 / 0.25} & \textbf{0.24 / 0.24} & 0.03 / 0.04 \\
\midrule

\multirow{4}{*}{32}
& PerspectiveFields & 16.32 / 16.32 & 16.15 / 16.15 & 7.22 / 7.22 & 15.07 / 15.07 & 12.63 / 12.63 & 8.70 / 8.70 \\
& FundamentalCalib  & 57.13 / 57.04 & 57.11 / 57.08 & 0.47 / 0.48  & 72.02 / 72.08  & 72.03 / 72.06  & 0.08 / 0.12 \\     
& DroidCalib        & 1.30 / 1.34   & 1.28 / 1.29   & 0.16 / 0.43 & 0.66 / 0.59   & 0.56 / 0.56   & 0.72 / 0.19 \\
& TrifocalCalib-Direct & 0.38 / 0.38 & 0.38 / 0.38 & 0.01 / 0.01 & \textbf{0.32 / 0.32} & \textbf{0.31 / 0.31} & 0.01 / 0.01 \\ 
& TrifocalCalib-MSAC & 0.66 / 0.66 & 0.66 / 0.66 & \textbf{2e-5 / 2e-5} & 0.37 / 0.37 & 0.37 / 0.37 & \textbf{3e-6 / 3e-6} \\
& TrifocalCalib-MSAC-Opt     & \textbf{0.21 / 0.21} & \textbf{0.21 / 0.21} & 0.001 / 0.001 & 0.40 / 0.40 & 0.40 / 0.40 & 0.004 / 0.004 \\
\bottomrule
\end{tabular}
\end{table*}

Table~\ref{tab:blendedmvs_results} summarizes the intrinsic calibration results on the BlendedMVS dataset~\cite{2020_blendedmvs}.
% for varying numbers of input images (8, 16, and 32) and different levels of synthetic noise applied to the initial intrinsics. Metrics reported are mean, median, and standard deviation of the relative error across $f_x$, $f_y$, $c_x$, and $c_y$, averaged over 100 runs per configuration.
% 
\textit{TrifocalCalib} consistently achieves the lowest error across all settings.
% , confirming its robustness and accuracy in image sequences.
%
\textit{TrifocalCalib-Direct}, despite its simplicity and lack of explicit outlier rejection, performs surprisingly well. It achieves the best results in half of the evaluated settings, which highlights the inherent robustness of Trifocal Tensor–based self-calibration.
\textit{TrifocalCalib-MSAC-Opt}, which combines robust model selection with further refinement via nonlinear optimization, consistently delivers low calibration error across noise levels and image counts and achieves the best results in the other half of the evaluated settings.
Even the purely selection-based \textit{TrifocalCalib-MSAC} achieves competitive performance with low variance, underscoring the effectiveness of MSAC-style initialization.
\textit{DroidCalib} performs moderately well for 16 or more images, but shows noticeably higher variance and reduced accuracy in the 8-image case.
\textit{FundamentalCalib} shows partly acceptable performance at low image counts but fails dramatically at 32 images. 
% producing extremely high errors.
% (e.g., 72\% mean error in Scene2). 
While this may seem counterintuitive, 
% - since more images typically provide more constraints - 
this result likely stems from the sequential image selection process. As views are added, the pairwise similarity scores decline, meaning later images are less overlapping and potentially introduce noisier or less reliable correspondences. 
% This setup amplifies sensitivity to outliers and noise, a known weakness of fundamental matrix–based self-calibration approaches as discussed in the related work. 
The result underscores that, despite the additional data, fundamental matrix–based methods are highly vulnerable to degraded match quality in large image sequences.
\textit{PerspectiveFields}, a single-image learning-based method, performs worst across all settings, with mean errors consistently above 13\% and high variance, which limits its usefulness in accurate self-calibration scenarios.

Overall, these results underline the strength of trifocal tensor constraints when coupled with robust initialization and optimization. The \textit{TrifocalCalib} variants provide a reliable and accurate solution for projective self-calibration, outperforming both classical and learning-based baselines in real data settings.

\subsection{Image Data - IAMCV Dataset}
\label{sec:evaluation_images_iamcv}

\begin{table}[t]
\centering
\caption{Intrinsic calibration error for the IAMCV dataset~\cite{2024-JKU-dataset}. Results are reported as mean in percent.}
% Lower is better. The absolute precision of the results is limited by the unknown ground truth quality; the purpose of this evaluation is to demonstrate that the algorithms remain effective in the practically relevant scenario of a forward-facing camera moving approximately on a plane.}
\label{tab:iamcv_results}
\begin{tabular}{l c}
\toprule
Method & Mean Calibration Error [\%] $\downarrow$ \\
\midrule
TrifocalCalib-Direct & 5.26 \\
TrifocalCalib-MSAC & \textbf{1.37} \\
TrifocalCalib-MSAC-Opt & 2.87 \\
DroidCalib & 2.88 \\
\bottomrule
\end{tabular}
\end{table}

Table~\ref{tab:iamcv_results} shows that TrifocalCalib-MSAC achieves the lowest mean calibration error, indicating strong robustness in this practical use case.
However, given the limited quality of the available ground truth, the absolute accuracy of the reported numbers should not be overinterpreted. Instead, this experiment serves to demonstrate that the proposed methods remain effective in the practically relevant scenario of a forward-facing camera moving approximately on a plane.
For results focused on absolute accuracy, we refer to our evaluation on the BlendedMVS dataset in Section~\ref{sec:evaluation_images}.

%\vspace{-5pt}
\section{Conclusion \& Future Work} 
\label{sec:conclusion}

% Summarize the main findings and contributions of the paper.
% Future research directions (e.g., improving robustness, testing on real-world datasets).

We presented \textit{TrifocalCalib}, a projective self-calibration method based on constraints from the calibrated trifocal tensor. The method jointly estimates focal length and principal point without relying on calibration targets or prior scene knowledge.
Evaluations on synthetic and real image data confirm that \textit{TrifocalCalib} outperforms classical methods as well as recent hybrid and deep learning-based approaches. Its robustness to noise, computational efficiency, and conceptual simplicity make it a compelling alternative for intrinsic camera calibration in uncontrolled environments.
Future work should focus on reporting confidence measures alongside calibration results, such as confidence intervals or uncertainty estimates, to better assess the reliability of parameter estimates in practical applications.

%\vspace{-5pt}

\section*{Acknowledgment}
This work was funded by the Austrian Research Promotion Agency (FFG), PDrive, project number: 901692.

We also thank the authors of~\cite{julia2017critical}, whose publicly available implementation accelerated the development of parts of our codebase.

%\addtolength{\textheight}{-12cm}   
% This command serves to balance the column lengths
% on the last page of the document manually. It shortens
% the textheight of the last page by a suitable amount.
% This command does not take effect until the next page
% so it should come on the page before the last. Make
% sure that you do not shorten the text height too much.

%%%%%%%%%%%%%%%%%%%%%%%%%%%%%%%%%%%%%%%%%%%%%%%%%%%%%%%%%%%%%%%%%%%%%%%%%%%%%%%%

%\section*{APPENDIX}
%Appendixes should appear before the acknowledgment.

%\section*{ACKNOWLEDGMENT}
%This work was supported by the Austrian Ministry for Climate Action, Environment, Energy, Mobility, Innovation and Technology (BMK) Endowed Professorship for Sustainable Transport Logistics 4.0., IAV France S.A.S.U., IAV GmbH, Austrian Post AG and the UAS Technikum Wien.

%%%%%%%%%%%%%%%%%%%%%%%%%%%%%%%%%%%%%%%%%%%%%%%%%%%%%%%%%%%%%%%%%%%%%%%%%%%%%%%%

\bibliographystyle{IEEEtran}
\bibliography{literature}

% \newpage
% \newpage
% \pagebreak

% \input{07_appendix}

\end{document}